\documentclass{opt2023} 

\usepackage{url}
\usepackage{booktabs}       
\usepackage{amsfonts}       
\usepackage{nicefrac}       
\usepackage{microtype}      
\usepackage{xcolor}         
\usepackage{amsmath}
\usepackage{amssymb}
\usepackage{mathtools}
\usepackage{float}
\usepackage[capitalize,noabbrev]{cleveref}
\usepackage{bbm}
\usepackage{appendix}
\usepackage{caption}
\usepackage{enumitem}
\usepackage{comment}

\title[Understanding the Role of Optimization in Double Descent]{Understanding the Role of Optimization in Double Descent}

\optauthor{%
\Name{Chris Yuhao Liu} \Email{yliu298@ucsc.edu}\\
\addr Department of Computer Science and Engineering \\
\addr University of California, Santa Cruz \\
\addr Santa Cruz, CA, USA
\AND
\Name{Jeffrey Flanigan} \Email{jmflanig@ucsc.edu}\\
\addr Department of Computer Science and Engineering \\
\addr University of California, Santa Cruz \\
\addr Santa Cruz, CA, USA
}

\begin{document}

\maketitle

\begin{abstract}%
The phenomenon of model-wise double descent, where the test error peaks and then reduces as the model size increases, is an interesting topic that has attracted the attention of researchers due to the striking observed gap between theory and practice \citep{Belkin2018ReconcilingMM}. Additionally, while double descent has been observed in various tasks and architectures, the peak of double descent can sometimes be noticeably absent or diminished, even without explicit regularization, such as weight decay and early stopping. In this paper, we investigate this intriguing phenomenon from the optimization perspective and propose a simple optimization-based explanation for why double descent sometimes occurs weakly or not at all. To the best of our knowledge, we are the first to demonstrate that many disparate factors contributing to model-wise double descent (initialization, normalization, batch size, learning rate, optimization algorithm) are unified from the viewpoint of optimization: model-wise double descent is observed if and only if the optimizer can find a sufficiently low-loss minimum. These factors directly affect the condition number of the optimization problem or the optimizer and thus affect the final minimum found by the optimizer, reducing or increasing the height of the double descent peak. We conduct a series of controlled experiments on random feature models and two-layer neural networks under various optimization settings, demonstrating this optimization-based unified view. Our results suggest the following implication: Double descent is unlikely to be a problem for real-world machine learning setups. Additionally, our results help explain the gap between weak double descent peaks in practice and strong peaks observable in carefully designed setups.
\end{abstract}


\section{Introduction}
\label{sec:introduction}

The phenomenon of double descent \citep{Belkin2018ReconcilingMM,Nakkiran2019DeepDD}, where the generalization error first decreases, increases, and then decreases again as the model size $P$ surpasses the dataset size $N$ (the interpolation threshold), has attracted a lot of attention from the machine learning community. This unexpected behavior calls into question the traditional understanding of generalization in both under- and over-parameterized regimes. Over the past few years, the double descent phenomenon has been observed in various models \citep{Nakkiran2019DeepDD,Chang2020ProvableBO,SahraeeArdakan2021AsymptoticsOR,Fonseca2022SimilarityAG} and learning paradigms \citep{Nakkiran2019DeepDD,Rice2020OverfittingIA,Hassani2022TheCO,Dar2020DoubleDD,Cotter2021DistillingDD}. Previous research has contributed to our understanding of the double descent phenomenon in various contexts and from different perspectives, including bias-variance trade-off tools \citep{Yang2020RethinkingBT,dAscoli2020DoubleTI}, VC theory \citep{Lee2022VCTE}, condition numbers \citep{Kuzborskij2021OnTR,Schaeffer2023DoubleDD} (the ratio between the maximum and the minimum singular values of the data matrix for linear regression), and aspects of optimization \citep{Kini2020AnalyticSO,Gamba2022DeepDD,Gamba2023OnTL}. Although double descent is ubiquitous across many machine learning and deep learning setups, it is not always observed \citep{Huang2020SelfAdaptiveTB,Dar2020SubspaceFM,Buschjger2021ThereIN,Ba2020GeneralizationOT,Wang2022OptimalAF}, and depends heavily on various factors \citep{Nakkiran2019DeepDD,Krijthe2016ThePP,dAscoli2020TripleDA,Lin2020WhatCT,Quetu2023CanWA,Quetu2023DodgingTS,Gu2023UnravelingTE}.

Many factors contribute to double descent occurring or not occurring in any given deep learning setup. These include the initialization, learning rate, scale of features, normalization, batch size, and choice of optimization algorithm. For the first time, we demonstrate that the effect of all these disparate factors is unified into a single phenomenon from the viewpoint of optimization: double descent is observed if and only if the given optimizer\footnote{We use the term ``optimizer'' to refer to an optimization algorithm configured with the corresponding hyperparameters.} can find a sufficiently low-loss minimum. For hyper-parameters affecting the underlying optimization problem, the condition number (the ratio between the largest and the smallest singular values of the feature matrix) is affected, which in turn affects the minimum being found by the optimizer. For hyper-parameters affecting the optimizer directly, those that lead to a worse minimum reduce the peak of double descent. Although, in hindsight, these results are intuitive and almost obvious, we are unaware of any prior work connecting these phenomena in this simple manner.

Our results on simple random feature models and two-layer neural networks indicate that classifiers in practice are unlikely to exhibit the peaking phenomenon. First, inductive biases and hyperparameters are usually chosen carefully using a validation set to prevent overfitting so that no model is over-trained. Second, even if a model overfits, many iterations are required for critically parameterized models to exhibit a strong double descent curve (\cref{sec:training_longer}). Therefore, realistic setups already act as a ``natural'' mitigator of double descent. While no existing theoretical framework explains all our results, we believe our results are particularly useful for suggesting avenues for further theoretical study.

\section{Related Work}
\label{sec:related_work}

There has been some previous work investigating when the double descent phenomena occur and when it doesn't \citep{Krijthe2016ThePP,Nakkiran2020OptimalRC,Huang2020SelfAdaptiveTB,Dar2020SubspaceFM,Buschjger2021ThereIN,Ba2020GeneralizationOT,Wang2022OptimalAF,Gu2023UnravelingTE}. Previous work has shown that the number of iterations (early-stopping) can affect the double descent peak, and sometimes eliminate it \citep{Nakkiran2019DeepDD,Bodin2021ModelSA,Yang2020RethinkingBT}. A recent study also argues that the double descent occurs due to label noise \citep{Gupta2023OnFS}, but we show that double descent can be observed with clean datasets. However, we are unaware of any work investigating the effect of normalization, learning rate, batch size, choice of optimization algorithm, or the other hyperparameters we investigate on the observed peak in double descent.

Apart from the double descent phenomenon in generalization error, prior work, from the optimization perspective, has also identified a similar descent-like phenomenon in the condition number for simple linear regression problems. Specifically, for least-squares linear regression, prior theoretical and experimental work has shown that the condition number increases near the peak of the double descent, where the \textit{condition number} here is the ratio between the maximum and the minimum singular values of the data matrix for linear regression. \citet{Poggio2019DoubleDI} is the first to show that the condition number for a random matrix peaks when the number of rows equals the number of columns for linear regression. \citet{Rangamani2020ForIK} demonstrated that the condition number regulates the stability of the least squares solution, which is why the error peaks at $P = N$. \citet{Kuzborskij2021OnTR} argued that the condition number of the feature matrix is determined only by the minimal singular value under the assumption that the maximum singular value is constant. Their theory included dependence of the excess risk on the minimal eigenvalue. \citet{Mei2019TheGE} hypothesized that the peaking at $P = N$ can be explained by the explosion of the variance, which is related to the condition number. 

\begin{figure}
    \centering
    \begin{minipage}{0.63\textwidth}
        \centering
        \includegraphics[width=\textwidth]{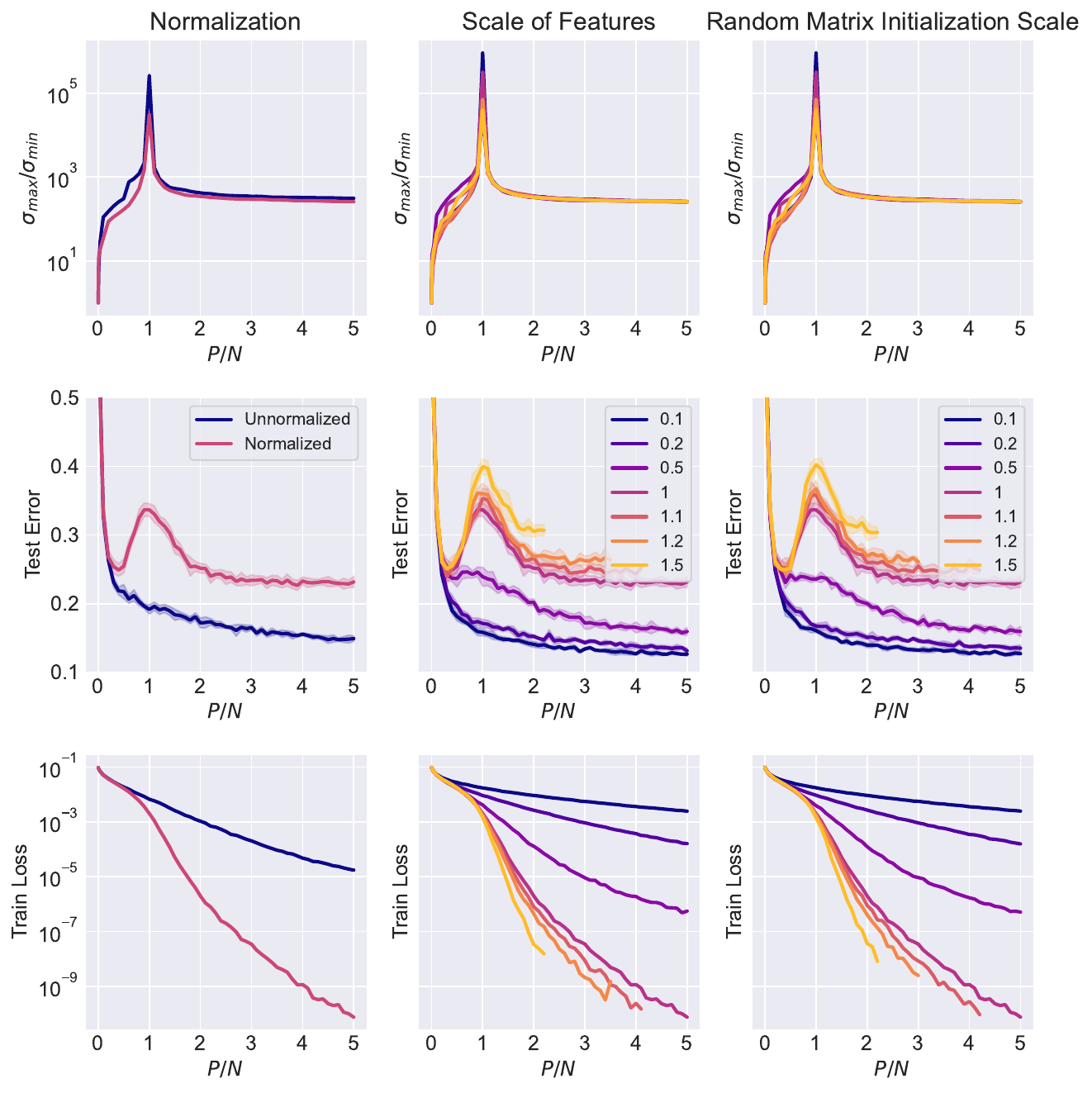}
    \end{minipage}%
    \hspace{1em}
    \begin{minipage}{0.32\textwidth}
        \centering
        \caption{\small Condition number, test error, and training loss on random feature models with varying properties features on an RFM with ReLU features. \textcolor{violet}{Darker colors} represent \textbf{higher condition number} in row 1, \textbf{weaker double descent} in row 2, and \textbf{high training loss} in row 3. \textbf{Row 1}: Normalizing the data and increasing the scale of the features and random matrix yield better (smaller) condition numbers at $P = N$. \textbf{Row 2}: Double descent does not occur in unnormalized, small-scale input features and small random matrix initialization. Double descent is observed more strongly when a lower condition number is at the peak. \textbf{Row 3}: A higher double descent peak corresponds to a lower training loss at $P = N$.}
        \label{fig:mnist_rfn_normalization_inputscale_initscale}
    \end{minipage}
\end{figure}

To the best of our knowledge, previous studies have theoretically identified the peak in the condition number of random matrices and observed it in some real datasets but have not established a connection to the double descent curve. Our work connects the previous finding on the peaking of the condition number and its effect on a double descent curve. In \cref{fig:mnist_rfn_normalization_inputscale_initscale}, the condition number is largest at $P=N$ (the double descent peak), which agrees with theoretical results. We find that this makes optimization harder at $P=N$, so models are less likely to converge, which causes the peak of double descent to be reduced or disappear, but reappear with longer training (\cref{sec:training_longer}). This finding extends the previous ones because we show the interplay of the condition numbers in features and other elements in optimization. Our observation on condition number also extends the previous work in that we can control the magnitude of double descent by controlling the condition number, as shown in \cref{fig:mnist_rfn_normalization_inputscale_initscale,fig:fashionmnist_rfn_normalization_inputscale_initscale,fig:cifar10_rfn_normalization_inputscale_initscale}. Our work is also related to the mitigation of double descent, which we briefly discuss in \cref{sec:double_decent_mitigation}.

\section{Experimental Setup}
\label{sec:experimental_setup}

We present empirical evidence on random feature models trained on MNIST \citep{lecun2010mnist} and include additional results for both random feature models and two-layer neural networks on Fashion-MNIST \citep{Xiao2017FashionMNISTAN} and CIFAR-10 \citep{Krizhevsky2009LearningML} in \cref{sec:additional_experiments}. Full experimental details are given in \cref{sec:experimental_setup_and_hyperparameters}. We also emphasize that we choose a setup that avoids known factors that could mitigate the double descent phenomenon as much as possible. Specifically, we do not use early stopping, which is known to remove the double descent phenomenon \citep{Nakkiran2019DeepDD,Bodin2021ModelSA,Yang2020RethinkingBT}. On the contrary, in the default setup, we train all models for a sufficient number of epochs (1000) so that most models converge in the first $1/10$ of the epochs. To further exclude the possibility of stopping too early, even in this scenario (which is not likely), we extend it to 10,000 epochs in \cref{sec:training_longer}. We also do not use any norm-based weight penalty to restrict the parameter space \citep{Nakkiran2020OptimalRC} or add dropout layers during training. See \cref{sec:experimental_setup_and_hyperparameters} for a detailed experimental setup.

\section{Poor Conditioning Reduces Double Descent}
\label{sec:poor_conditioning}

It is well-known from optimization theory that better conditioning leads to faster convergence for gradient descent optimization \citep{Nesterov2018LecturesOC}. We observe that the height of the double descent peak negatively correlates with the condition number of the random feature matrices. In \cref{fig:mnist_rfn_normalization_inputscale_initscale}, we see that, at $P = N$, the peak in condition number is always present, regardless of the peak in double descent. However, the height of the condition number matters because the setting with a lower condition number (lighter color in row 1) corresponds to a more prominent double descent peak (lighter color in row 2). Here, the poor conditioning of the optimization problem makes it more difficult for the model around $P/N = 1$ and the optimization algorithm to converge to a sufficiently low-loss minimum, as illustrated in the third row in \cref{fig:mnist_rfn_normalization_inputscale_initscale}. Previous studies also have shown the condition number peaks at the peak of the double descent (which occurs at $P=N$) \citep{Poggio2019DoubleDI,Rangamani2020ForIK,Huang2020LargeSA,Kuzborskij2021OnTR,Chen2021ConditioningOR}. We observe this to be the case in \cref{fig:mnist_rfn_normalization_inputscale_initscale}, confirming the findings of these previous studies.

We see that the double descent peaks disappear when the feature matrix is 1) \textbf{unnormalized}, or when the random matrix/features have a smaller scale due to the 2) \textbf{scaling of the features} (e.g., scaling the feature matrix by a small constant) or 3) \textbf{the initialization of the random matrix} (e.g., from a normal distribution with a small variance) (because they all change the input features to the linear classifier). We plot the training loss and observe that the setting where double descent occurs has a training loss much smaller than one in which double descent does not occur (\cref{fig:mnist_rfn_normalization_inputscale_initscale}, bottom).
The condition number is largest at $P=N$ (the double descent peak) in \cref{fig:mnist_rfn_normalization_inputscale_initscale}, which agrees with prior theoretical results (see \cref{sec:related_work}). We find that this makes optimization harder at $P=N$, so models do not converge, which causes the peak of double descent to be reduced or disappear, but reappear with longer training (\cref{sec:training_longer}). We observe that double descent tends to occur in better-conditioned matrices because a lower minimum is found by the optimizer.

\section{Slow-Convergence Leads to Disappearance of Double Descent}
\label{sec:slow_convergence}

Our results suggest that a slow-convergence setting\footnote{We use the term ``slow-convergence setting'' to refer to situations where the model converges slowly given the optimization problem, the optimization algorithm, its associated hyperparameters, and all other hyperparameters.} often reduces or removes the peaking phenomenon, and fast-convergence setting (i.e., finding a lower minimum) restores the peaking phenomenon. We observe this pattern in hyperparameters that affect the optimizer, such as \textbf{learning rate} (constant and decay), \textbf{batch size}, and \textbf{optimization algorithm}. All three factors can be considered as affecting the convergence of the optimizer directly, where a higher minimum found by the optimizer corresponds to a less prominent peak.

\begin{figure}
    \centering
    \includegraphics[width=\textwidth]{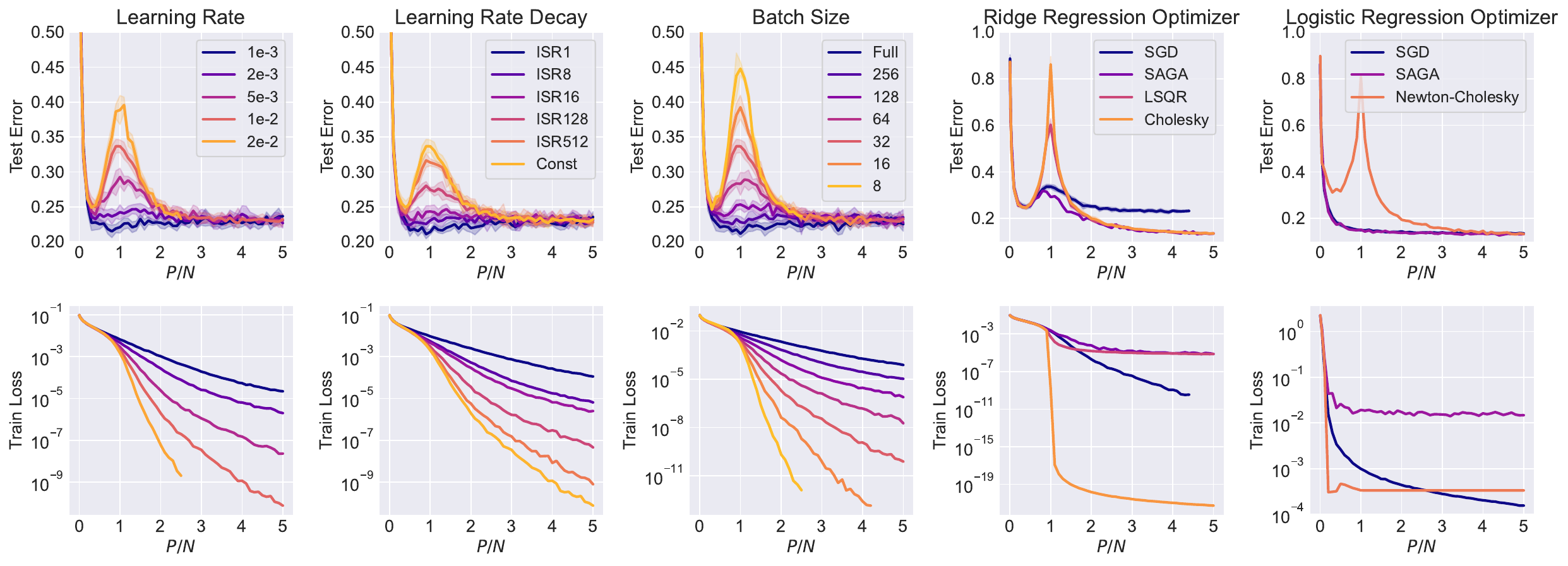}
    \caption{Test error and training loss on random feature models with varying learning rates, batch sizes, and optimization algorithms. Double descent occurs with a sufficiently large and steady learning rate, a small enough batch size, or a low-enough minimum is obtained by the optimizer.}
    \label{fig:mnist_rfn_lr_lrsche_bs_opt}
\end{figure}

We observe that a faster-convergence optimization algorithm that finds a lower loss minimum exhibits double descent more strongly, and slow-convergence settings may not exhibit it at all (last two columns of \cref{fig:mnist_rfn_lr_lrsche_bs_opt}).
We observe that the peak height negatively correlates with the training loss at $P = N$. For curves with Cholesky decomposition, the error rate approaches random guessing in both ridge regression and logistic regression. Similar observations can be made on SAGA and SGD, where the double descent peak occurs mildly with a slight increase in test error.

We observe that low learning rates, which are insufficient to reach a low-loss minimum, reduce or eliminate the double descent peak (first column of \cref{fig:mnist_rfn_lr_lrsche_bs_opt}).  When the overparameterization ratio $P/N$ is greater than $3$, using a lower learning rate has almost no impact on the test error, but it eliminates the peak without any form of explicit regularization. The same result holds for learning rate decay. We observe that faster learning rate decay has a similar effect to a small constant learning rate, and decaying every iteration removes the peak entirely. Both imply that the emergence of double descent requires maintaining a stable and large enough learning rate during training, corresponding to a faster convergence optimization setting that lands at a lower minimum loss.

We observe that large batch sizes reduce or eliminate the double descent phenomenon. In \cref{fig:mnist_rfn_lr_lrsche_bs_opt}, the peaking phenomenon disappears on the generalization curve for batch sizes of 500 (full-batch) and 256. Modifying the batch size changes the number of updates the optimization algorithm performs. Thus, large batch sizes take fewer steps.

\section{Training Longer Recovers Double Descent}
\label{sec:training_longer}

\begin{figure}
    \centering
    \includegraphics[width=\textwidth]{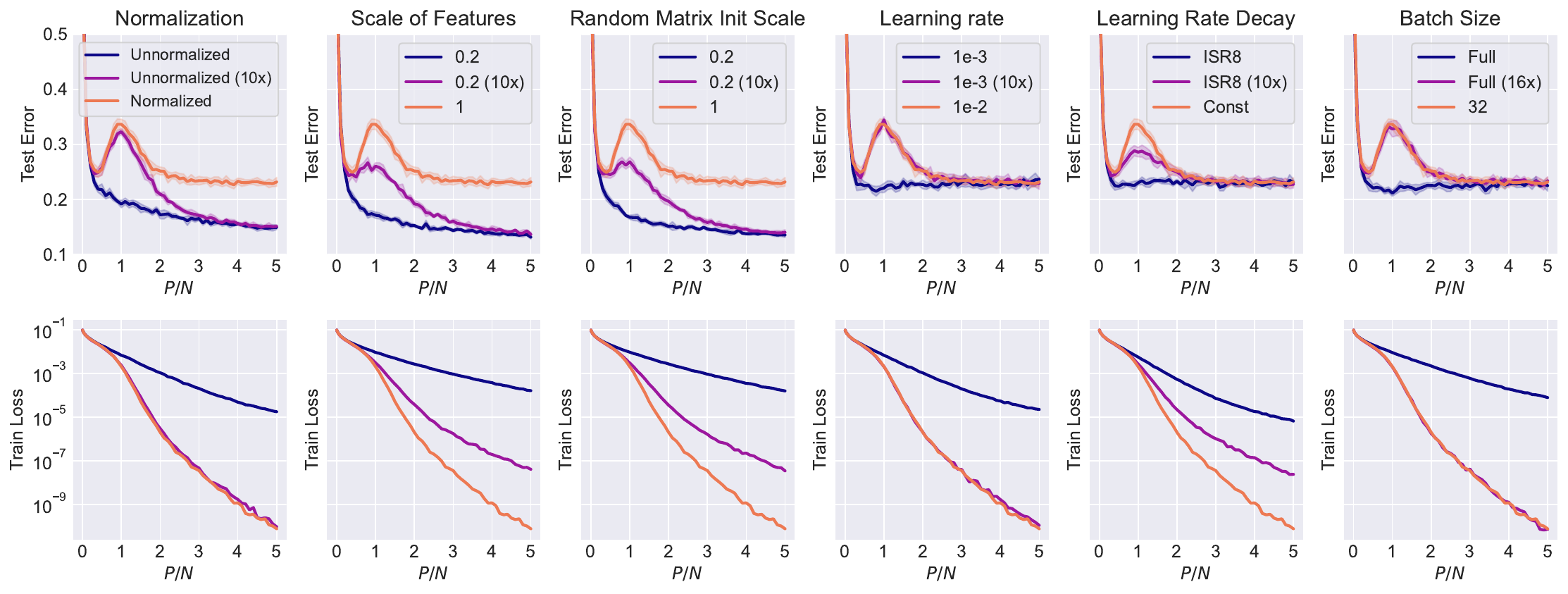}
    \caption{Test error and training loss on random feature models with varying learning rates, batch sizes, and optimization algorithms at 10x iterations. The peaking phenomenon is recovered as long as a sufficient number of gradient updates is applied, even for ill-conditioned features.}
    \label{fig:mnist_rfn_poor_optimizers_10x}
\end{figure}

We show that for hyper-parameter setups that do not exhibit double descent, we can recover this phenomenon simply by running the optimization procedure longer to reach a lower training loss. In \cref{fig:mnist_rfn_poor_optimizers_10x}, we increase the number of iterations by a factor of 10 so that the training loss of a slow-convergence setting is aligned with or approaches the default (a much lower one). We observe that the double descent peak is recovered for all factors. Precisely, in \cref{fig:mnist_rfn_poor_optimizers_10x} for the learning rate (column 4) and batch size (column 6), both the test error and training loss curves with ten times the original epochs exactly overlap with curves produced by the faster-converging optimization setup (i.e., the larger learning rate of $1e^{-2}$ or the smaller batch size of $32$). Our results suggest that the optimization length is a simple but strong indicator of why double descent is observed in some realistic settings.

In practical settings, proper regularization often prevents models from reaching 0 training error. Even though no explicit regularization is applied, we usually do not continue training for a large number of epochs when the model has already overfit the training set. In our experiments, for double descent to recover \cref{fig:mnist_rfn_poor_optimizers_10x}, models are usually trained 200-400 times longer after converging to 0 training error, which is not realistic for deep and large models used in practice. This also implies that the peak's emergence rate largely depends on the specific optimization setup, where optimization settings that converge slowly do not exhibit the peaking behavior.

\section{Experiments with Two-Layer Neural Networks}
\label{sec:two_layer_nns}

We present results on two-layer neural networks in \cref{fig:mnist_mlp} in the Appendix. Despite the fact that two-layer neural networks are fully non-linear models, we observe that the results on two-layer neural networks are consistent with our previous findings on random feature models.

\section{Conclusion}
\label{sec:conclusion}

In dissecting the occurrence of double descent in machine learning, our study elucidates a unifying underlying phenomenon tied to optimization: double descent occurs when the optimizer can achieve a low-loss minimum. While seemingly disconnected, factors like initialization, learning rate, normalization, batch size, and optimizer choice work together to influence the overall optimization trajectory, thereby affecting the optimizer's path to a minimum. Our findings not only simplify the understanding of these variables but also shed light on the practical likelihood of double descent, which is minimized due to careful hyperparameter selections and inherent inductive biases in real-world setups. Although current theoretical frameworks fall short of encapsulating our observations entirely, our results shed light on promising directions for deeper theoretical scrutiny into the interplay between optimization and double descent, thereby paving the way for a deeper understanding of the intriguing phenomenon.

\bibliography{references}

\begin{thebibliography}{57}
\providecommand{\natexlab}[1]{#1}
\providecommand{\url}[1]{\texttt{#1}}
\expandafter\ifx\csname urlstyle\endcsname\relax
  \providecommand{\doi}[1]{doi: #1}\else
  \providecommand{\doi}{doi: \begingroup \urlstyle{rm}\Url}\fi

\bibitem[Ba et~al.(2020)Ba, Erdogdu, Suzuki, Wu, and Zhang]{Ba2020GeneralizationOT}
Jimmy Ba, Murat~A. Erdogdu, Taiji Suzuki, Denny Wu, and Tianzong Zhang.
\newblock Generalization of two-layer neural networks: An asymptotic viewpoint.
\newblock In \emph{8th International Conference on Learning Representations, {ICLR} 2020, Addis Ababa, Ethiopia, April 26-30, 2020}. OpenReview.net, 2020.
\newblock URL \url{https://openreview.net/forum?id=H1gBsgBYwH}.

\bibitem[Belkin et~al.(2018)Belkin, Hsu, Ma, and Mandal]{Belkin2018ReconcilingMM}
Mikhail Belkin, Daniel~J. Hsu, Siyuan Ma, and Soumik Mandal.
\newblock Reconciling modern machine-learning practice and the classical bias-variance trade-off.
\newblock \emph{Proceedings of the National Academy of Sciences}, 116:\penalty0 15849--15854, 2018.

\bibitem[Bodin and Macris(2021)]{Bodin2021ModelSA}
Anthony Bodin and Nicolas Macris.
\newblock Model, sample, and epoch-wise descents: exact solution of gradient flow in the random feature model.
\newblock In \emph{Neural Information Processing Systems}, 2021.
\newblock URL \url{https://api.semanticscholar.org/CorpusID:239616354}.

\bibitem[Buschj{\"a}ger and Morik(2021)]{Buschjger2021ThereIN}
Sebastian Buschj{\"a}ger and Katharina Morik.
\newblock There is no double-descent in random forests.
\newblock \emph{ArXiv preprint}, abs/2111.04409, 2021.
\newblock URL \url{https://arxiv.org/abs/2111.04409}.

\bibitem[Chang et~al.(2020)Chang, Li, Oymak, and Thrampoulidis]{Chang2020ProvableBO}
Xiangyu Chang, Yingcong Li, Samet Oymak, and Christos Thrampoulidis.
\newblock Provable benefits of overparameterization in model compression: From double descent to pruning neural networks.
\newblock In \emph{AAAI Conference on Artificial Intelligence}, 2020.

\bibitem[Chen et~al.(2021)Chen, Wang, and Kyrillidis]{Chen2021MitigatingDD}
John Chen, Qihan Wang, and Anastasios Kyrillidis.
\newblock Mitigating deep double descent by concatenating inputs.
\newblock \emph{Proceedings of the 30th ACM International Conference on Information \& Knowledge Management}, 2021.

\bibitem[Chen and Schaeffer(2021)]{Chen2021ConditioningOR}
Zhijun Chen and Hayden Schaeffer.
\newblock Conditioning of random feature matrices: Double descent and generalization error.
\newblock \emph{ArXiv preprint}, abs/2110.11477, 2021.
\newblock URL \url{https://arxiv.org/abs/2110.11477}.

\bibitem[Cotter et~al.(2021)Cotter, Menon, Narasimhan, Rawat, Reddi, and Zhou]{Cotter2021DistillingDD}
Andrew Cotter, Aditya~Krishna Menon, Harikrishna Narasimhan, Ankit~Singh Rawat, Sashank~J. Reddi, and Yichen Zhou.
\newblock Distilling double descent.
\newblock \emph{ArXiv preprint}, abs/2102.06849, 2021.
\newblock URL \url{https://arxiv.org/abs/2102.06849}.

\bibitem[Dar and Baraniuk(2020)]{Dar2020DoubleDD}
Yehuda Dar and Richard Baraniuk.
\newblock Double double descent: On generalization errors in transfer learning between linear regression tasks.
\newblock \emph{ArXiv preprint}, abs/2006.07002, 2020.
\newblock URL \url{https://arxiv.org/abs/2006.07002}.

\bibitem[Dar et~al.(2020)Dar, Mayer, Luzi, and Baraniuk]{Dar2020SubspaceFM}
Yehuda Dar, Paul Mayer, Lorenzo Luzi, and Richard~G. Baraniuk.
\newblock Subspace fitting meets regression: The effects of supervision and orthonormality constraints on double descent of generalization errors.
\newblock In \emph{Proceedings of the 37th International Conference on Machine Learning, {ICML} 2020, 13-18 July 2020, Virtual Event}, volume 119 of \emph{Proceedings of Machine Learning Research}, pages 2366--2375. {PMLR}, 2020.
\newblock URL \url{http://proceedings.mlr.press/v119/dar20a.html}.

\bibitem[d'Ascoli et~al.(2020{\natexlab{a}})d'Ascoli, Refinetti, Biroli, and Krzakala]{dAscoli2020DoubleTI}
St{\'e}phane d'Ascoli, Maria Refinetti, Giulio Biroli, and Florent Krzakala.
\newblock Double trouble in double descent : Bias and variance(s) in the lazy regime.
\newblock In \emph{International Conference on Machine Learning}, 2020{\natexlab{a}}.

\bibitem[d'Ascoli et~al.(2020{\natexlab{b}})d'Ascoli, Sagun, and Biroli]{dAscoli2020TripleDA}
St{\'e}phane d'Ascoli, Levent Sagun, and Giulio Biroli.
\newblock Triple descent and the two kinds of overfitting: where and why do they appear?
\newblock \emph{Journal of Statistical Mechanics: Theory and Experiment}, 2021, 2020{\natexlab{b}}.

\bibitem[Defazio et~al.(2014)Defazio, Bach, and Lacoste{-}Julien]{Defazio2014SAGAAF}
Aaron Defazio, Francis~R. Bach, and Simon Lacoste{-}Julien.
\newblock {SAGA:} {A} fast incremental gradient method with support for non-strongly convex composite objectives.
\newblock In Zoubin Ghahramani, Max Welling, Corinna Cortes, Neil~D. Lawrence, and Kilian~Q. Weinberger, editors, \emph{Advances in Neural Information Processing Systems 27: Annual Conference on Neural Information Processing Systems 2014, December 8-13 2014, Montreal, Quebec, Canada}, pages 1646--1654, 2014.
\newblock URL \url{https://proceedings.neurips.cc/paper/2014/hash/ede7e2b6d13a41ddf9f4bdef84fdc737-Abstract.html}.

\bibitem[Dhifallah and Lu(2020)]{Dhifallah2020APP}
Oussama Dhifallah and Yue~M. Lu.
\newblock A precise performance analysis of learning with random features.
\newblock \emph{ArXiv preprint}, abs/2008.11904, 2020.
\newblock URL \url{https://arxiv.org/abs/2008.11904}.

\bibitem[Emami et~al.(2020)Emami, Sahraee{-}Ardakan, Pandit, Rangan, and Fletcher]{Emami2020GeneralizationEO}
Melikasadat Emami, Mojtaba Sahraee{-}Ardakan, Parthe Pandit, Sundeep Rangan, and Alyson~K. Fletcher.
\newblock Generalization error of generalized linear models in high dimensions.
\newblock In \emph{Proceedings of the 37th International Conference on Machine Learning, {ICML} 2020, 13-18 July 2020, Virtual Event}, volume 119 of \emph{Proceedings of Machine Learning Research}, pages 2892--2901. {PMLR}, 2020.
\newblock URL \url{http://proceedings.mlr.press/v119/emami20a.html}.

\bibitem[Fonseca and Guidetti(2022)]{Fonseca2022SimilarityAG}
Nayara Fonseca and Veronica Guidetti.
\newblock Similarity and generalization: From noise to corruption.
\newblock \emph{ArXiv preprint}, abs/2201.12803, 2022.
\newblock URL \url{https://arxiv.org/abs/2201.12803}.

\bibitem[Gamba et~al.(2022)Gamba, Englesson, Bjorkman, and Azizpour]{Gamba2022DeepDD}
Matteo Gamba, Erik Englesson, Marten Bjorkman, and Hossein Azizpour.
\newblock Deep double descent via smooth interpolation.
\newblock \emph{ArXiv preprint}, abs/2209.10080, 2022.
\newblock URL \url{https://arxiv.org/abs/2209.10080}.

\bibitem[Gamba et~al.(2023)Gamba, Azizpour, and Bjorkman]{Gamba2023OnTL}
Matteo Gamba, Hossein Azizpour, and Marten Bjorkman.
\newblock On the lipschitz constant of deep networks and double descent.
\newblock \emph{ArXiv preprint}, abs/2301.12309, 2023.
\newblock URL \url{https://arxiv.org/abs/2301.12309}.

\bibitem[Gu et~al.(2023)Gu, Zheng, and Aste]{Gu2023UnravelingTE}
Yufei Gu, Xiaoqing Zheng, and Tomaso Aste.
\newblock Unraveling the enigma of double descent: An in-depth analysis through the lens of learned feature space.
\newblock \emph{ArXiv preprint}, abs/2310.13572, 2023.
\newblock URL \url{https://arxiv.org/abs/2310.13572}.

\bibitem[Gupta et~al.(2023)Gupta, Mishra, Luu, and Bouassami]{Gupta2023OnFS}
Arunav Gupta, Rohit Mishra, William Luu, and Mehdi Bouassami.
\newblock On feature scaling of recursive feature machines.
\newblock \emph{ArXiv preprint}, abs/2303.15745, 2023.
\newblock URL \url{https://arxiv.org/abs/2303.15745}.

\bibitem[Hassani and Javanmard(2022)]{Hassani2022TheCO}
Hamed Hassani and Adel Javanmard.
\newblock The curse of overparametrization in adversarial training: Precise analysis of robust generalization for random features regression.
\newblock \emph{ArXiv preprint}, abs/2201.05149, 2022.
\newblock URL \url{https://arxiv.org/abs/2201.05149}.

\bibitem[Hastie et~al.(2019)Hastie, Montanari, Rosset, and Tibshirani]{Hastie2019SurprisesIH}
Trevor~J. Hastie, Andrea Montanari, Saharon Rosset, and Ryan~J. Tibshirani.
\newblock Surprises in high-dimensional ridgeless least squares interpolation.
\newblock \emph{Annals of statistics}, 50 2:\penalty0 949--986, 2019.

\bibitem[Huang and Yang(2020)]{Huang2020LargeSA}
Hanwen Huang and Qinglong Yang.
\newblock Large scale analysis of generalization error in learning using margin based classification methods.
\newblock \emph{Journal of Statistical Mechanics: Theory and Experiment}, 2020, 2020.

\bibitem[Huang et~al.(2020{\natexlab{a}})Huang, Zhang, and Zhang]{Huang2020SelfAdaptiveTB}
Lang Huang, Chao Zhang, and Hongyang Zhang.
\newblock Self-adaptive training: beyond empirical risk minimization.
\newblock In Hugo Larochelle, Marc'Aurelio Ranzato, Raia Hadsell, Maria{-}Florina Balcan, and Hsuan{-}Tien Lin, editors, \emph{Advances in Neural Information Processing Systems 33: Annual Conference on Neural Information Processing Systems 2020, NeurIPS 2020, December 6-12, 2020, virtual}, 2020{\natexlab{a}}.
\newblock URL \url{https://proceedings.neurips.cc/paper/2020/hash/e0ab531ec312161511493b002f9be2ee-Abstract.html}.

\bibitem[Huang et~al.(2021)Huang, Zhang, and Zhang]{Huang2021SelfAdaptiveTB}
Lang Huang, Chaoning Zhang, and Hongyang Zhang.
\newblock Self-adaptive training: Bridging the supervised and self-supervised learning.
\newblock \emph{IEEE transactions on pattern analysis and machine intelligence}, PP, 2021.

\bibitem[Huang et~al.(2020{\natexlab{b}})Huang, Hogg, and Villar]{Huang2020DimensionalityRR}
Ningyuan~Teresa Huang, David~W. Hogg, and Soledad Villar.
\newblock Dimensionality reduction, regularization, and generalization in overparameterized regressions.
\newblock \emph{SIAM J. Math. Data Sci.}, 4:\penalty0 126--152, 2020{\natexlab{b}}.

\bibitem[Hui and Belkin(2020)]{Hui2020EvaluationON}
Like Hui and Mikhail Belkin.
\newblock Evaluation of neural architectures trained with square loss vs cross-entropy in classification tasks.
\newblock \emph{ArXiv preprint}, abs/2006.07322, 2020.
\newblock URL \url{https://arxiv.org/abs/2006.07322}.

\bibitem[Kan et~al.(2020)Kan, Nagy, and Ruthotto]{Kan2020AvoidingTD}
Kelvin~K. Kan, James~G. Nagy, and Lars Ruthotto.
\newblock Avoiding the double descent phenomenon of random feature models using hybrid regularization.
\newblock \emph{ArXiv preprint}, abs/2012.06667, 2020.
\newblock URL \url{https://arxiv.org/abs/2012.06667}.

\bibitem[Kini and Thrampoulidis(2020)]{Kini2020AnalyticSO}
Ganesh~Ramachandra Kini and Christos Thrampoulidis.
\newblock Analytic study of double descent in binary classification: The impact of loss.
\newblock \emph{2020 IEEE International Symposium on Information Theory (ISIT)}, pages 2527--2532, 2020.

\bibitem[Krijthe and Loog(2016)]{Krijthe2016ThePP}
Jesse~H. Krijthe and M.~Loog.
\newblock The peaking phenomenon in semi-supervised learning.
\newblock In \emph{International Workshop on Structural and Syntactic Pattern Recognition}, 2016.

\bibitem[Krizhevsky(2009)]{Krizhevsky2009LearningML}
Alex Krizhevsky.
\newblock Learning multiple layers of features from tiny images.
\newblock Technical report, 2009.

\bibitem[Kuzborskij et~al.(2021)Kuzborskij, Szepesvari, Rivasplata, Rannen-Triki, and Pascanu]{Kuzborskij2021OnTR}
Ilja Kuzborskij, Csaba Szepesvari, Omar Rivasplata, Amal Rannen-Triki, and Razvan Pascanu.
\newblock On the role of optimization in double descent: A least squares study.
\newblock In \emph{Neural Information Processing Systems}, 2021.

\bibitem[LeCun et~al.(2010)LeCun, Cortes, and Burges]{lecun2010mnist}
Yann LeCun, Corinna Cortes, and CJ~Burges.
\newblock Mnist handwritten digit database.
\newblock \emph{ATT Labs [Online]. Available: http://yann.lecun.com/exdb/mnist}, 2, 2010.

\bibitem[Lee and Cherkassky(2022)]{Lee2022VCTE}
Eng~Hock Lee and Vladimir Cherkassky.
\newblock Vc theoretical explanation of double descent.
\newblock \emph{ArXiv preprint}, abs/2205.15549, 2022.
\newblock URL \url{https://arxiv.org/abs/2205.15549}.

\bibitem[Lin and Dobriban(2020)]{Lin2020WhatCT}
Licong Lin and Edgar Dobriban.
\newblock What causes the test error? going beyond bias-variance via anova.
\newblock \emph{J. Mach. Learn. Res.}, 22:\penalty0 155:1--155:82, 2020.

\bibitem[Loureiro et~al.(2022)Loureiro, Gerbelot, Refinetti, Sicuro, and Krzakala]{Loureiro2022FluctuationsBV}
Bruno Loureiro, C'edric Gerbelot, Maria Refinetti, Gabriele Sicuro, and Florent Krzakala.
\newblock Fluctuations, bias, variance \& ensemble of learners: Exact asymptotics for convex losses in high-dimension.
\newblock \emph{ArXiv preprint}, abs/2201.13383, 2022.
\newblock URL \url{https://arxiv.org/abs/2201.13383}.

\bibitem[Luzi et~al.(2021)Luzi, Dar, and Baraniuk]{Luzi2021DoubleDA}
Lorenzo Luzi, Yehuda Dar, and Richard Baraniuk.
\newblock Double descent and other interpolation phenomena in gans.
\newblock \emph{ArXiv preprint}, abs/2106.04003, 2021.
\newblock URL \url{https://arxiv.org/abs/2106.04003}.

\bibitem[Mei and Montanari(2019)]{Mei2019TheGE}
Song Mei and Andrea Montanari.
\newblock The generalization error of random features regression: Precise asymptotics and the double descent curve.
\newblock \emph{Communications on Pure and Applied Mathematics}, 75, 2019.

\bibitem[Misra(2020)]{Misra2020MishAS}
Diganta Misra.
\newblock Mish: {A} self regularized non-monotonic activation function.
\newblock In \emph{31st British Machine Vision Conference 2020, {BMVC} 2020, Virtual Event, UK, September 7-10, 2020}. {BMVA} Press, 2020.
\newblock URL \url{https://www.bmvc2020-conference.com/assets/papers/0928.pdf}.

\bibitem[Nakkiran et~al.(2020{\natexlab{a}})Nakkiran, Kaplun, Bansal, Yang, Barak, and Sutskever]{Nakkiran2019DeepDD}
Preetum Nakkiran, Gal Kaplun, Yamini Bansal, Tristan Yang, Boaz Barak, and Ilya Sutskever.
\newblock Deep double descent: Where bigger models and more data hurt.
\newblock In \emph{8th International Conference on Learning Representations, {ICLR} 2020, Addis Ababa, Ethiopia, April 26-30, 2020}. OpenReview.net, 2020{\natexlab{a}}.
\newblock URL \url{https://openreview.net/forum?id=B1g5sA4twr}.

\bibitem[Nakkiran et~al.(2020{\natexlab{b}})Nakkiran, Venkat, Kakade, and Ma]{Nakkiran2020OptimalRC}
Preetum Nakkiran, Prayaag Venkat, Sham~M. Kakade, and Tengyu Ma.
\newblock Optimal regularization can mitigate double descent.
\newblock \emph{ArXiv preprint}, abs/2003.01897, 2020{\natexlab{b}}.
\newblock URL \url{https://arxiv.org/abs/2003.01897}.

\bibitem[Nesterov(2018)]{Nesterov2018LecturesOC}
Yurii Nesterov.
\newblock Lectures on convex optimization.
\newblock 2018.
\newblock URL \url{https://api.semanticscholar.org/CorpusID:125291746}.

\bibitem[Paige and Saunders(1982)]{Paige1982LSQRAA}
Christopher~C. Paige and Michael~A. Saunders.
\newblock Lsqr: An algorithm for sparse linear equations and sparse least squares.
\newblock \emph{ACM Trans. Math. Softw.}, 8:\penalty0 43--71, 1982.

\bibitem[Patil et~al.(2022{\natexlab{a}})Patil, Du, and Kuchibhotla]{Patil2022BaggingIO}
Pratik~V. Patil, Jin-Hong Du, and Arun~K. Kuchibhotla.
\newblock Bagging in overparameterized learning: Risk characterization and risk monotonization.
\newblock 2022{\natexlab{a}}.

\bibitem[Patil et~al.(2022{\natexlab{b}})Patil, Kuchibhotla, Wei, and Rinaldo]{Patil2022MitigatingMD}
Pratik~V. Patil, Arun~K. Kuchibhotla, Yuting Wei, and Alessandro Rinaldo.
\newblock Mitigating multiple descents: A model-agnostic framework for risk monotonization.
\newblock \emph{ArXiv preprint}, abs/2205.12937, 2022{\natexlab{b}}.
\newblock URL \url{https://arxiv.org/abs/2205.12937}.

\bibitem[Poggio et~al.(2019)Poggio, Kur, and Banburski]{Poggio2019DoubleDI}
Tomaso~A. Poggio, Gil Kur, and Andy Banburski.
\newblock Double descent in the condition number.
\newblock \emph{ArXiv preprint}, abs/1912.06190, 2019.
\newblock URL \url{https://arxiv.org/abs/1912.06190}.

\bibitem[Qu'etu and Tartaglione(2023{\natexlab{a}})]{Quetu2023CanWA}
Victor Qu'etu and Enzo Tartaglione.
\newblock Can we avoid double descent in deep neural networks?
\newblock 2023{\natexlab{a}}.

\bibitem[Qu'etu and Tartaglione(2023{\natexlab{b}})]{Quetu2023DodgingTS}
Victor Qu'etu and Enzo Tartaglione.
\newblock Dodging the sparse double descent.
\newblock \emph{ArXiv preprint}, abs/2303.01213, 2023{\natexlab{b}}.
\newblock URL \url{https://arxiv.org/abs/2303.01213}.

\bibitem[Rangamani et~al.(2020)Rangamani, Rosasco, and Poggio]{Rangamani2020ForIK}
Akshay Rangamani, Lorenzo Rosasco, and Tomaso~A. Poggio.
\newblock For interpolating kernel machines, minimizing the norm of the erm solution maximizes stability.
\newblock \emph{Analysis and Applications}, 2020.

\bibitem[Rice et~al.(2020)Rice, Wong, and Kolter]{Rice2020OverfittingIA}
Leslie Rice, Eric Wong, and J.~Zico Kolter.
\newblock Overfitting in adversarially robust deep learning.
\newblock In \emph{Proceedings of the 37th International Conference on Machine Learning, {ICML} 2020, 13-18 July 2020, Virtual Event}, volume 119 of \emph{Proceedings of Machine Learning Research}, pages 8093--8104. {PMLR}, 2020.
\newblock URL \url{http://proceedings.mlr.press/v119/rice20a.html}.

\bibitem[Sahraee-Ardakan et~al.(2021)Sahraee-Ardakan, Mai, Rao, Rossi, Rangan, and Fletcher]{SahraeeArdakan2021AsymptoticsOR}
Mojtaba Sahraee-Ardakan, Tung Mai, Anup~B. Rao, Ryan~A. Rossi, Sundeep Rangan, and Alyson~K. Fletcher.
\newblock Asymptotics of ridge regression in convolutional models.
\newblock In \emph{International Conference on Machine Learning}, 2021.

\bibitem[Schaeffer et~al.(2023)Schaeffer, Khona, Robertson, Boopathy, Pistunova, Rocks, Fiete, and Koyejo]{Schaeffer2023DoubleDD}
Rylan Schaeffer, Mikail Khona, Zachary Robertson, Akhilan Boopathy, Kateryna Pistunova, Jason~W. Rocks, Ila~Rani Fiete, and Oluwasanmi Koyejo.
\newblock Double descent demystified: Identifying, interpreting \& ablating the sources of a deep learning puzzle.
\newblock \emph{ArXiv preprint}, abs/2303.14151, 2023.
\newblock URL \url{https://arxiv.org/abs/2303.14151}.

\bibitem[Singla et~al.(2021)Singla, Singla, Jacobs, and Feizi]{Singla2021LowCA}
Vasu Singla, Sahil Singla, David Jacobs, and Soheil Feizi.
\newblock Low curvature activations reduce overfitting in adversarial training.
\newblock \emph{2021 IEEE/CVF International Conference on Computer Vision (ICCV)}, pages 16403--16413, 2021.

\bibitem[Wang and Bento(2022)]{Wang2022OptimalAF}
Jianxin Wang and Jos{\'e} Bento.
\newblock Optimal activation functions for the random features regression model.
\newblock \emph{ArXiv preprint}, abs/2206.01332, 2022.
\newblock URL \url{https://arxiv.org/abs/2206.01332}.

\bibitem[Wilson and Izmailov(2020)]{Wilson2020BayesianDL}
Andrew~Gordon Wilson and Pavel Izmailov.
\newblock Bayesian deep learning and a probabilistic perspective of generalization.
\newblock \emph{ArXiv preprint}, abs/2002.08791, 2020.
\newblock URL \url{https://arxiv.org/abs/2002.08791}.

\bibitem[Xiao et~al.(2017)Xiao, Rasul, and Vollgraf]{Xiao2017FashionMNISTAN}
Han Xiao, Kashif Rasul, and Roland Vollgraf.
\newblock Fashion-mnist: a novel image dataset for benchmarking machine learning algorithms.
\newblock \emph{ArXiv preprint}, abs/1708.07747, 2017.
\newblock URL \url{https://arxiv.org/abs/1708.07747}.

\bibitem[Yang et~al.(2020)Yang, Yu, You, Steinhardt, and Ma]{Yang2020RethinkingBT}
Zitong Yang, Yaodong Yu, Chong You, Jacob Steinhardt, and Yi~Ma.
\newblock Rethinking bias-variance trade-off for generalization of neural networks.
\newblock \emph{ArXiv preprint}, abs/2002.11328, 2020.
\newblock URL \url{https://arxiv.org/abs/2002.11328}.

\end{thebibliography}

\newpage

\appendix

\section{Experimental Setup and Hyperparameters}
\label{sec:experimental_setup_and_hyperparameters}

In this section, we described the detailed setup of the dataset, models, and training.

\subsection{Datasets}

We perform all experiments on MNIST \citep{lecun2010mnist}, Fashion-MNIST \citep{Xiao2017FashionMNISTAN}, and CIFAR-10 \citep{Krizhevsky2009LearningML}. We select small subsets of size $N$ from the full training set as our training data and evaluate the generalization error using the complete test set. We trained models for RFMs with a random feature size of up to $P / N = 5$, while for two-layer NNs, we utilize models up to $P / N = 5 \cdot C$ parameters, where $C$ is fixed at 10 in our setting. For two-layer NNs, the interpolation threshold is at $N \cdot C$ instead of $N$ \citep{Belkin2018ReconcilingMM}. By default, we normalize the image pixel features to follow a normal distribution by applying the transformation $\frac{\mathbf{X} - \mu}{s} \cdot \gamma$, where $\mu, s, \gamma$ are the mean, standard deviation, and the scaling factor, respectively.

\subsection{Training}

\paragraph{Models.}
For random feature models, we use static first-layer weights $\overline{\mathbf{W}}_0 \in \mathbb{R}^{d \times P}$ and trainable second-layer weights $\mathbf{W}_1 \in \mathbb{R}^{P \times C}$, where $P$ and $C$ represent the projected feature dimension and the number of classes, respectively. We initialize both weight matrices from $\overline{\mathbf{W}}_0 \sim \mathcal{N} \left( 0, \frac{k_0}{\sqrt{D}} \right)$ and $\mathbf{W}_1 \sim \mathcal{N} \left( 0, \frac{k_1}{\sqrt{P}} \right)$, where $k_0$ is a scaling factor for the standard deviation of the weight matrix, and $D$ represents the input feature dimension. $k_1$ is always 1 in all experiments. Bias terms are always set to 0, and the ReLU nonlinearity is used by default. Our setup for two-layer neural networks resembles that of a random feature model. The only difference is that the first layer weights are trained.

\paragraph{Optimization.} In our learning rate decay experiments, we adopt an inverse square root schedule similar to \citet{Nakkiran2019DeepDD}, but we multiply the initial learning rate by the factor $\frac{1}{\sqrt{\lfloor t/l \rfloor}+1}$, where $t$ is the current iteration and $l$ is an interval parameter. By controlling the interval $l$, we modify the decay frequency during the training trajectory. In optimizer experiments, we select numerical solvers, Cholesky and QR \citep{Paige1982LSQRAA}, for ridge regression and Newton-Cholesky for logistic regression because they obtain solutions with much lower loss than SGD. We also chose SAGA \citep{Defazio2014SAGAAF} as an alternative gradient-based algorithm that (empirically) converges slower than SGD with momentum for comparison. A small regularization constant 1e-8 is used for numerical stability. In the 10x iteration experiments, we ensure that the number of gradient updates matches that in a mini-batch case. Given a full-batch setting, we calculate the extended number of iterations by $\mathcal{T}_{\text{new}} = \lceil N / b \rceil \cdot \mathcal{T}$, which is equal to the number of gradient updates in a mini-batch setting with batch size $b$ and $\mathcal{T}$ iterations.

\paragraph{Producing double descents.} Consistent with \citet{Belkin2018ReconcilingMM} and \citet{Nakkiran2019DeepDD}, we employed the same set of hyperparameters for all model sizes and trained them using SGD with a fixed number of epochs and constant step size. The default hyperparameters are selected based on two training error constraints: 1) the largest model has to attain 0 training error within the first $1/10$ iterations, and 2) (at least) all models with $P \geq N$ have to converge to 0 training error before the final iteration. These constraints, derived from empirical observations, are fast in convergence and effective in generating the double descent phenomenon for both RFM and two-layer NN trained with SGD on MNIST and Fashion-MNIST. After some exploration, we use SGD with a Nesterov momentum of 0.95, a mini-batch size of 32, and a constant step size of 1e-2 for 1000 epochs. For two-layer NNs, we increase the step size to 5e-2 and the number of epochs to 1500. By default, we utilize the standard MSE loss. When a specific hyperparameter is studied, all other parameters follow the default ones. We primarily focus on the MSE loss because it has been heavily studied in the literature \citep{Belkin2018ReconcilingMM,Hastie2019SurprisesIH,Mei2019TheGE,Emami2020GeneralizationEO,Dhifallah2020APP}. We follow the previous implementation and do not include softmax at the network output \citet{Belkin2018ReconcilingMM,Hui2020EvaluationON}. In our experiments, we strictly control the confounding variables with potential effects on the peak in all experiments, such as label noise, which could exacerbate the curve \citep{Nakkiran2019DeepDD,Gu2023UnravelingTE}, and all forms of weight decay and early stopping, which are known to flatten the peak \citep{Nakkiran2020OptimalRC}. This ensures the robustness of our results. All experiments are repeated at least five times.

\section{Related Work on Double Descent Mitigation}
\label{sec:double_decent_mitigation}

While we do not intend to mitigate the double descent, our work investigates conditions under which the peak of double descent does or does not occur. Related, various techniques have been demonstrated to successfully reduce the peak, including $\ell_2$ regularization \citep{Mei2019TheGE,Nakkiran2019DeepDD,Nakkiran2020OptimalRC,dAscoli2020TripleDA,Lin2020WhatCT,Kan2020AvoidingTD,Quetu2023CanWA,Quetu2023DodgingTS}, ensemble methods \citep{Wilson2020BayesianDL,dAscoli2020TripleDA,Loureiro2022FluctuationsBV,Patil2022BaggingIO}, cross-validation \citep{Patil2022MitigatingMD}, dimensionality reduction \citep{Huang2020DimensionalityRR}, input concatenation \citep{Chen2021MitigatingDD}, and the type of non-linear random features \citep{Dhifallah2020APP}. Several works claim that double descent is not observed in certain settings, even without the explicit mitigation techniques mentioned above. These settings include self-adaptive training \citep{Huang2020SelfAdaptiveTB,Huang2021SelfAdaptiveTB}, level of supervision \citep{Dar2020SubspaceFM,Luzi2021DoubleDA}, random forest models \citep{Buschjger2021ThereIN}, a two-layer neural network with certain initialization of the first layer weight \citep{Ba2020GeneralizationOT}, and special activation functions \citep{Wang2022OptimalAF,Singla2021LowCA}. We highlight that our focus in this paper is not to propose a technique to mitigate double descent. Instead, we find that the conditioning of the optimization problem and specific setups significantly affects the magnitude of the peaking phenomenon, and models in slower-convergence settings do not exhibit the peak. However, this observation might be helpful as a simple technique to mitigate double descent in practice, which we leave to future research. We also emphasize the importance of carefully examining the effect of optimization in producing double descent. This is because optimization is involved in almost all settings mentioned above. Therefore, it is crucial to identify whether the absence of double descent is due to optimization before examining other more complicated factors.

\section{Additional Experiments}
\label{sec:additional_experiments}

Here, we present the exact figures as in the paper's main text but on additional datasets (Fashion-MNIST and CIFAR-10) and two-layer neural networks. Our findings in the paper's main text generalize to these datasets and models.

\subsection{Time Evolution of A Double Descent Curve}
\label{sec:time_evolution}

\begin{figure}
    \centering
    \includegraphics[width=0.8\textwidth]{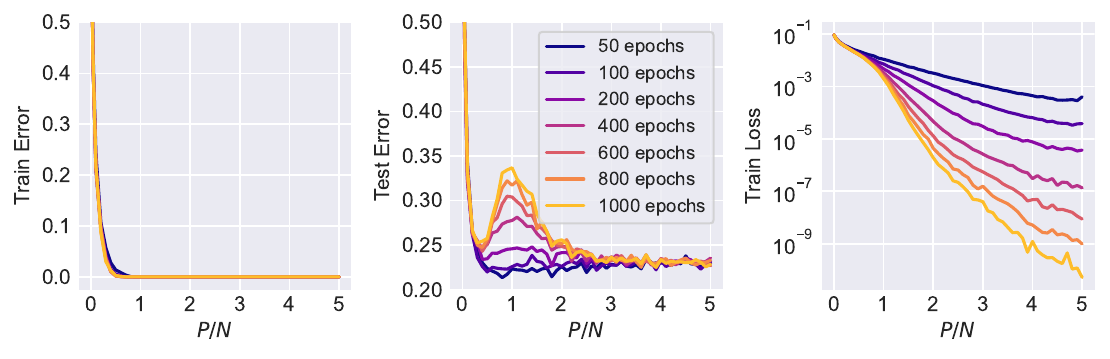}
    \caption{The time evolution of a double descent curve. At the interpolation threshold $P = N$, all models have achieved zero training error, and the peak starts to emerge only after zero training error is obtained.}
    \label{fig:time_evolution}
\end{figure}

We present the temporal evolution of a double descent curve through gradient descent iterations, demonstrating that the peaks form only post-model convergence. This aligns with the theoretical findings and synthetic experiments by \citet{Bodin2021ModelSA}, indicating that the peaking phenomenon at $P \approx N$ appears only beyond a certain iteration, with a monotonic curve prevailing prior. Figure \cref{fig:time_evolution} illustrates the iteration-wise change in training/test error and loss. We discern that models at $P \approx N$ converge on the training set after 50 epochs. Moreover, a training span of fewer than 200 epochs is inadequate for these models to manifest the peaking phenomenon. Despite the model maintaining a zero training error up to the 1000-th epoch, the peak in test error initiates and escalates as the training loss goes down. This insight corroborates preceding studies \citep{Nakkiran2019DeepDD,Bodin2021ModelSA}, asserting that early stopping diminishes the peak. Yet, we contend that the use of early stopping in hyperparameter tuning might be excessive for this goal, given double descent necessitates a substantial number of iterations post-convergence to appear. Hence, from an optimization standpoint, mitigating double descent is feasible, provided a model is not over-trained.

\subsection{Ill-Conditioned Non-Linear Features}

\begin{figure}
    \centering
    \subfigure[Varying nonlinearities]{\includegraphics[width=0.4\textwidth]{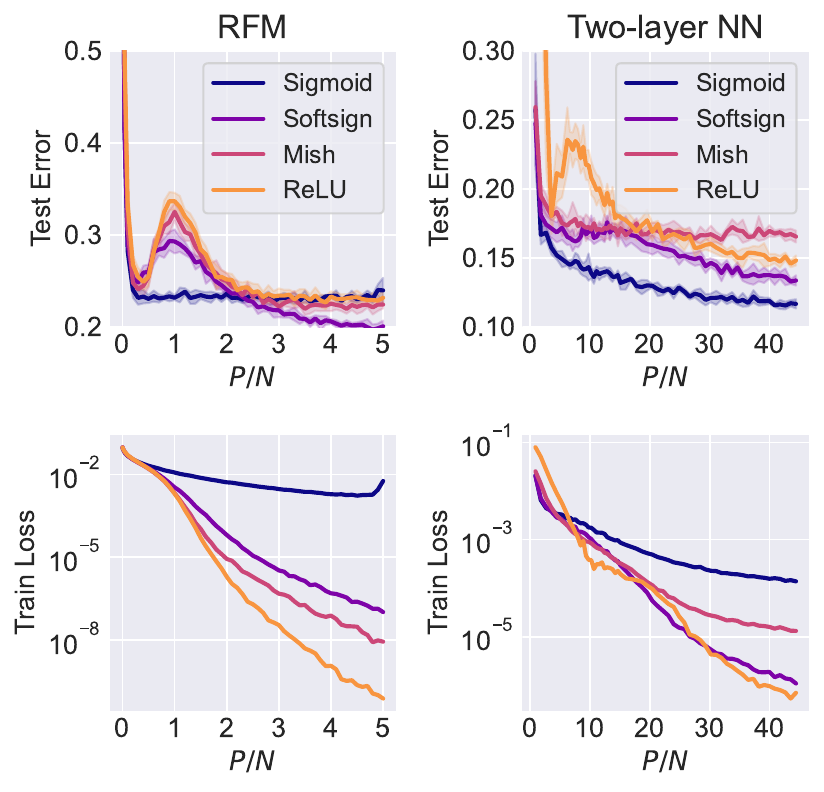}}
    \subfigure[Increasing iterations]{\includegraphics[width=0.4\textwidth]{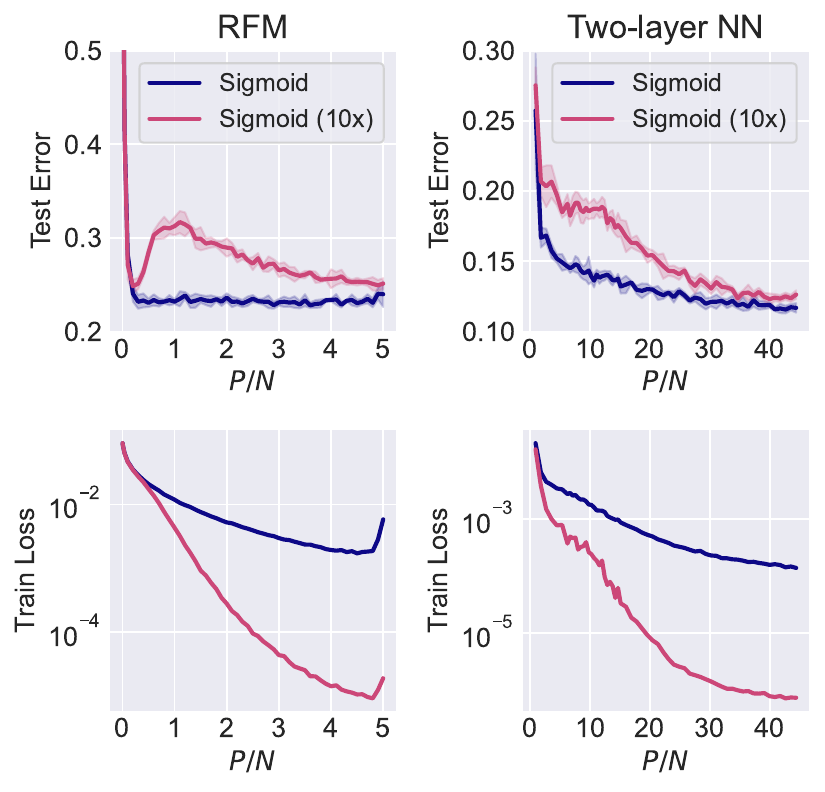}}
    \caption{\textbf{Left}: Test error and training loss curves for RFMs trained and four different activation functions. Models with activation functions that converge a higher training loss tend to avoid double descent. Given the same optimization and hyperparameter setup, the sigmoid activation does not exhibit double descent in RFM and two-layer neural networks. \textbf{Right}: Test error and training loss curves for RFMs trained and sigmoid activation. 10x iterations recover double descent. This matches our findings on the impact of a slow-convergence setting on double descent.}
    \label{fig:mnist_activation}
\end{figure}

For random feature models, another way to modify the input features is the choice of the non-linear function. Previous work shows that activation functions reduce the peaking phenomenon \citep{Dhifallah2020APP}, but we show that we can recover the peak by simply increasing the number of iterations. In the left figure of \cref{fig:mnist_activation}, we employ ReLU, mish \citep{Misra2020MishAS}, softsign, and sigmoid nonlinearities. ReLU and sigmoid show consistent behavior for both RFM and two-layer neural networks, even though activation functions operate differently on these two models (i.e., one with the input and the other with the intermediate embeddings). However, for mish and softsign, double descent is only observed in RFMs. In the second figure of \cref{fig:mnist_activation}, we show that the absence of double descent on sigmoid follows the same pattern as our previous experiments in \cref{fig:mnist_rfn_poor_optimizers_10x}. Non-linear features produced by sigmoid make optimization difficult, resulting in monotonicity, but we can recover the peaking phenomenon by scaling the number of iterations by 10.

\begin{figure}[H]
    \centering
    \includegraphics[width=0.8\textwidth]{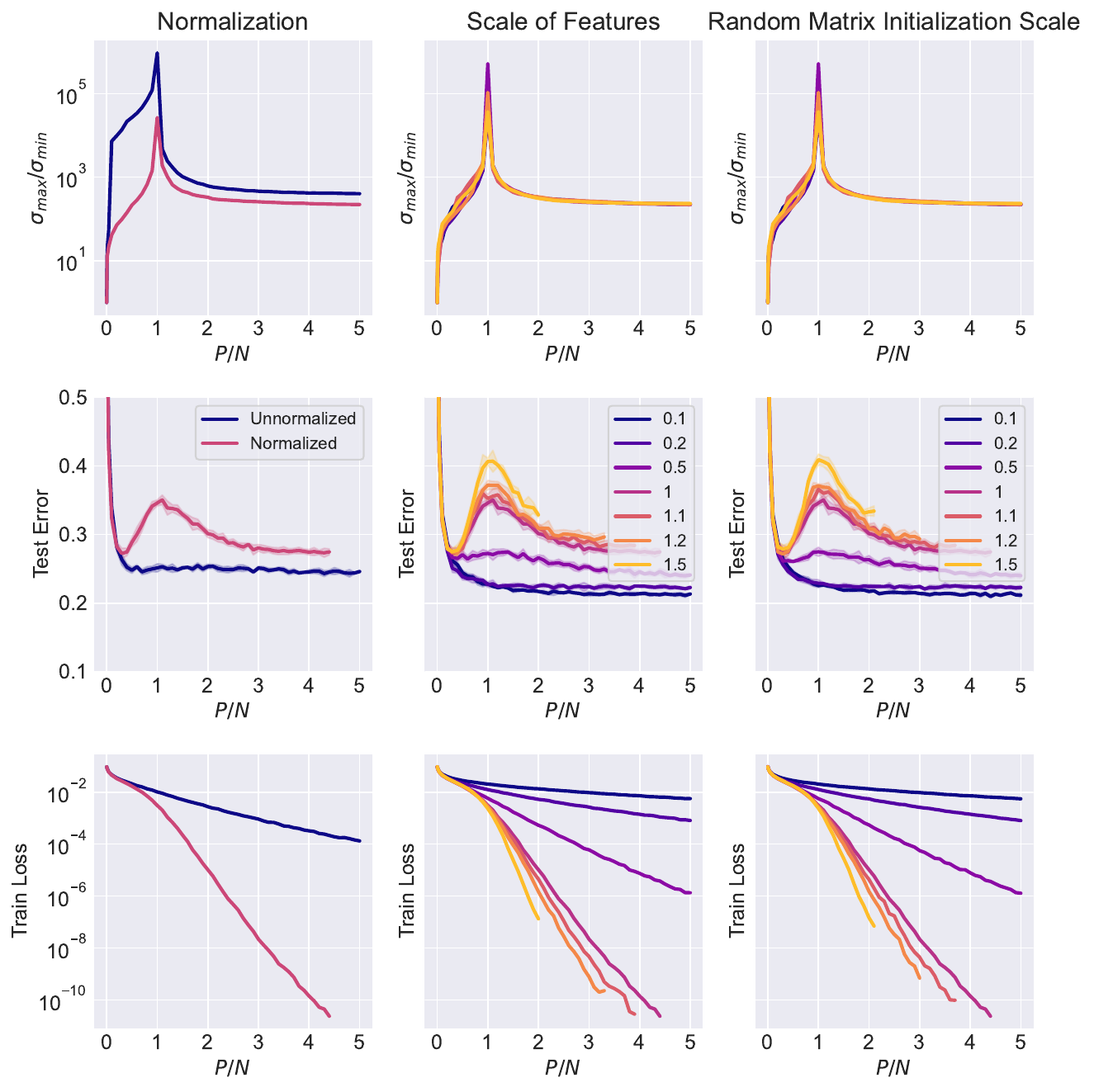}
    \caption{Additional results to support \cref{fig:mnist_rfn_normalization_inputscale_initscale} using the Fashion-MNIST dataset.}
    \label{fig:fashionmnist_rfn_normalization_inputscale_initscale}
\end{figure}

\begin{figure}[H]
    \centering
    \includegraphics[width=0.8\textwidth]{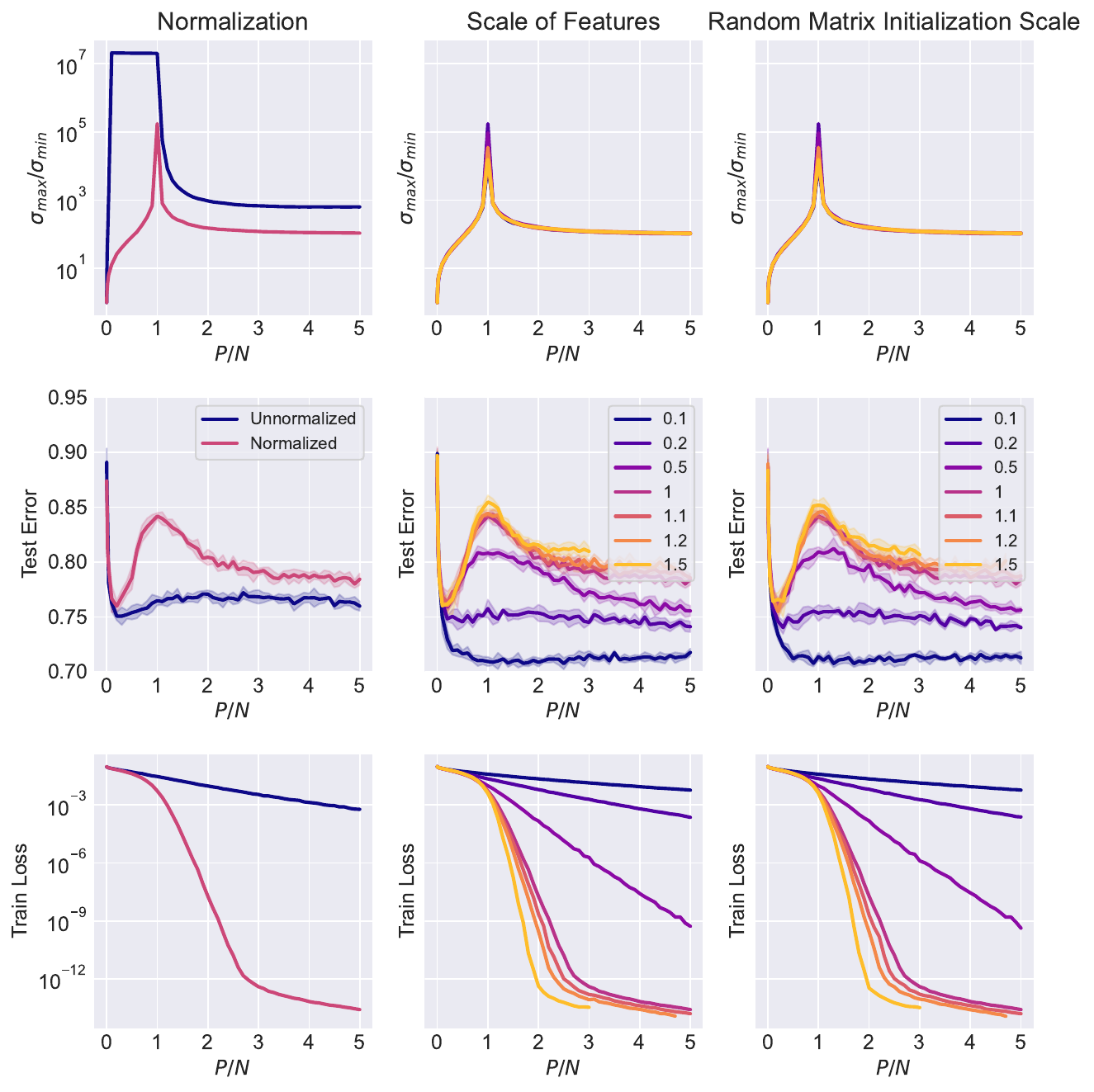}
    \caption{Additional results to support \cref{fig:mnist_rfn_normalization_inputscale_initscale} using the CIFAR-10 dataset.}
    \label{fig:cifar10_rfn_normalization_inputscale_initscale}
\end{figure}

\begin{figure}[H]
    \centering
    \subfigure[Two-layer neural network on MNIST]{\includegraphics[width=\textwidth]{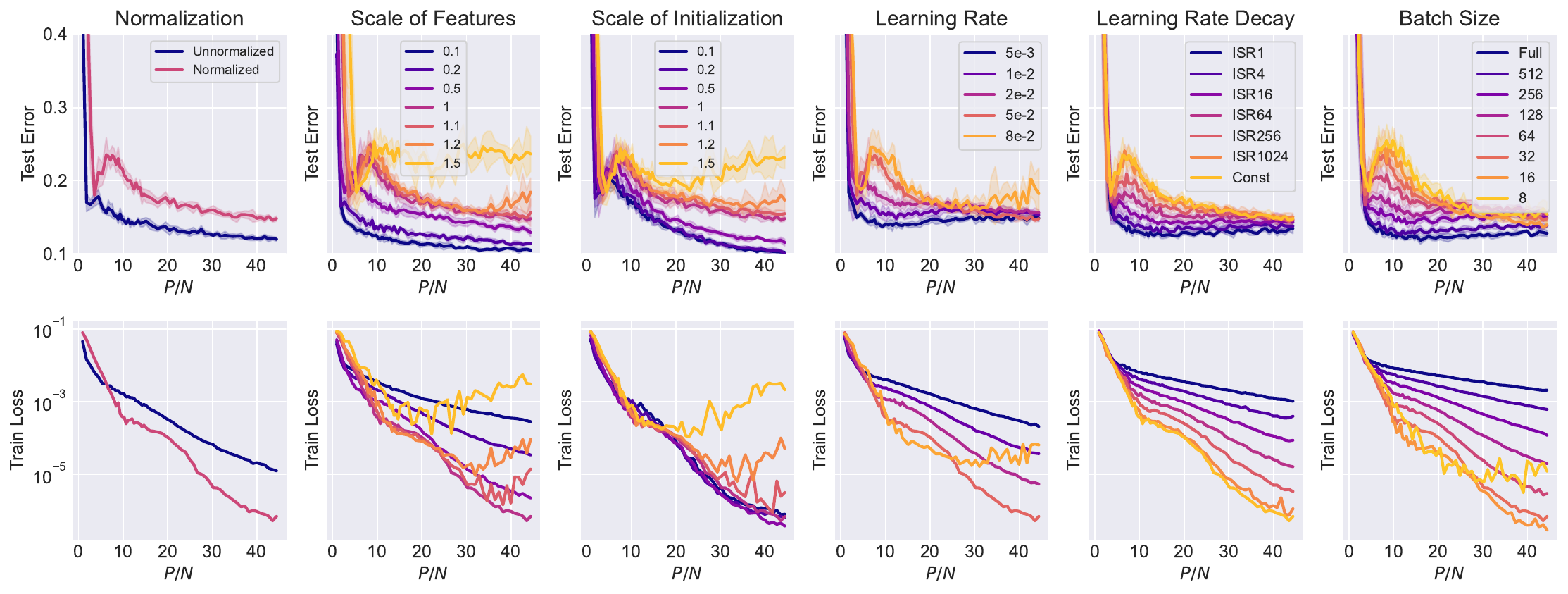}}
    \subfigure[Two-layer neural network on Fashion-MNIST]{\includegraphics[width=\textwidth]{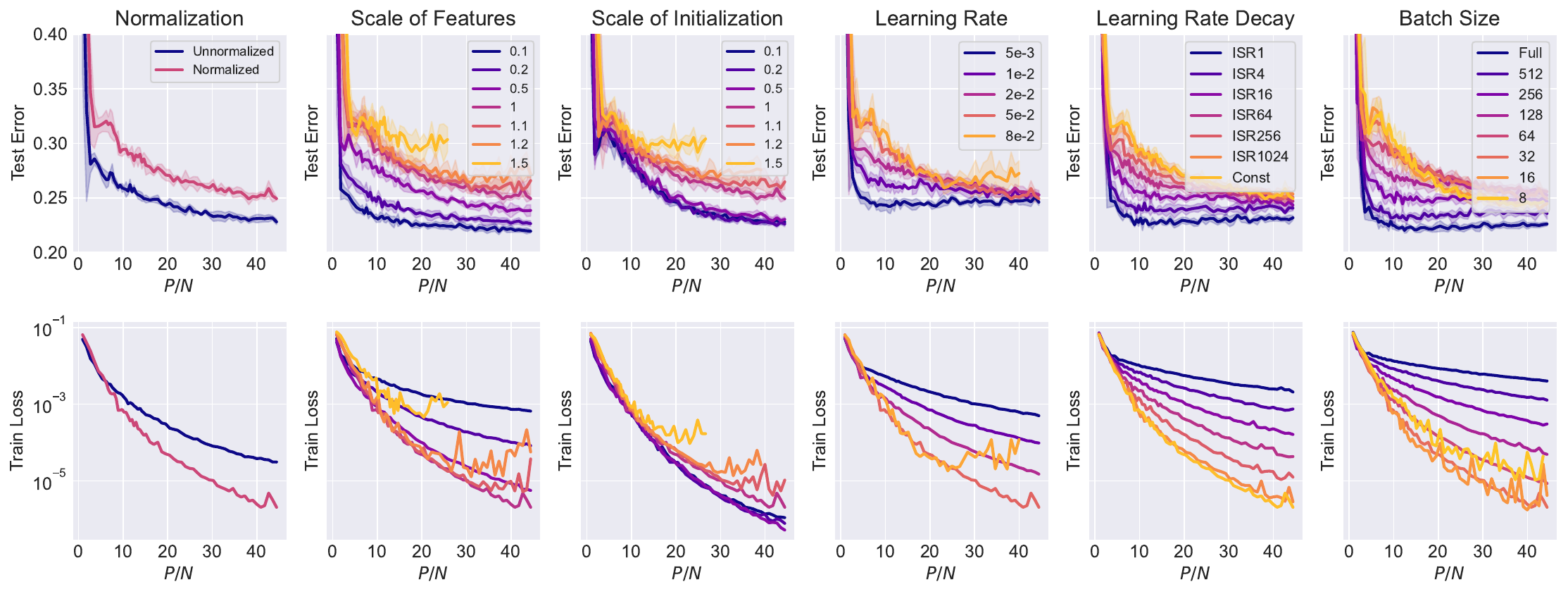}}
    \caption{Additional results to support \cref{fig:mnist_rfn_normalization_inputscale_initscale,fig:mnist_rfn_lr_lrsche_bs_opt}. Test error and training loss curves by (using) normalization, varying scale of features or initialization, learning rate, batch size, and optimization algorithm of a two-layer neural network on MNIST and Fashion-MNIST. The peaking phenomenon becomes less prominent as the features become worse-conditioned or directly using a slow-convergence setting.}
\end{figure}

\subsection{Slow-Convergence Settings in Optimization}

\begin{figure}[H]
    \centering
    \subfigure[RFM on Fashion-MNIST]{\includegraphics[width=\textwidth]{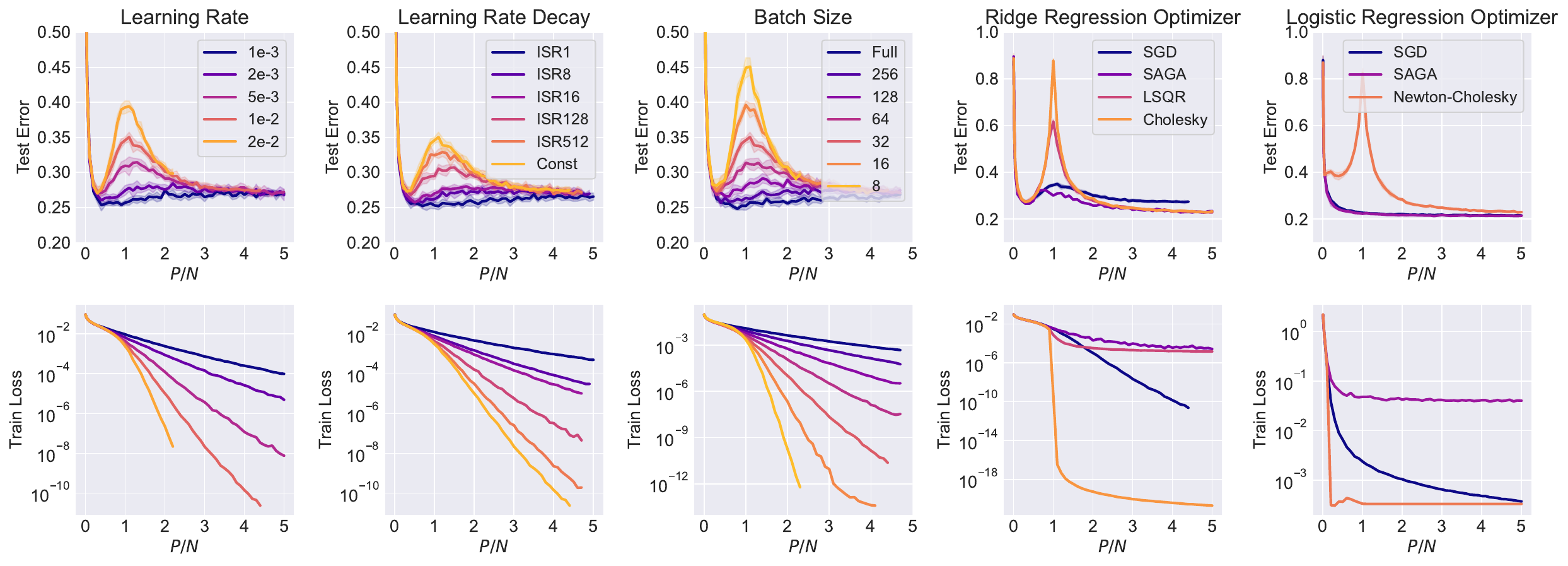}}
    \subfigure[RFM on CIFAR-10]{\includegraphics[width=\textwidth]{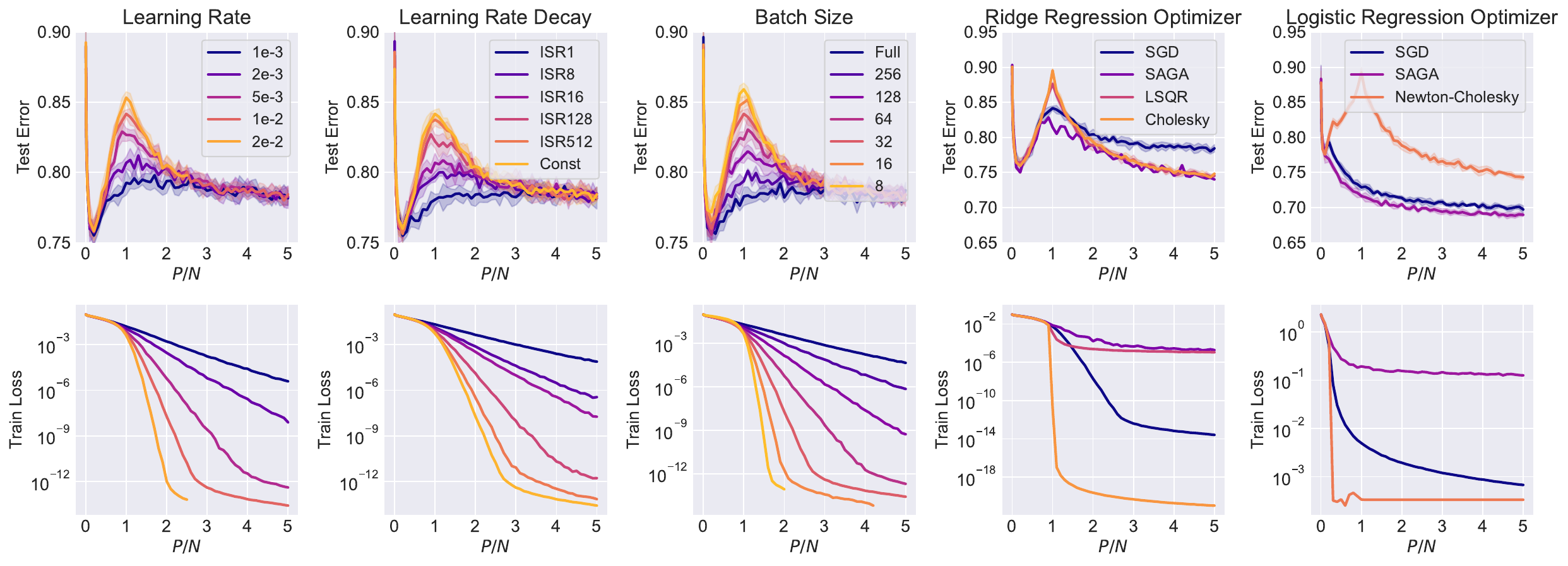}}
    \caption{Additional results to support \cref{fig:mnist_rfn_lr_lrsche_bs_opt}. Test error and training loss curves by varying learning rate, batch size, and optimization algorithm of a random feature model on Fashion-MNIST and CIFAR-10. The peaks are reduced on slow-convergence settings with too small a learning rate, too frequent learning rate decay, too large batch size, and slow-convergence settings.}
    \label{fig:mnist_mlp}
\end{figure}

\begin{figure}[H]
    \centering
    \subfigure[RFM on Fashion-MNIST]{\includegraphics[width=\textwidth]{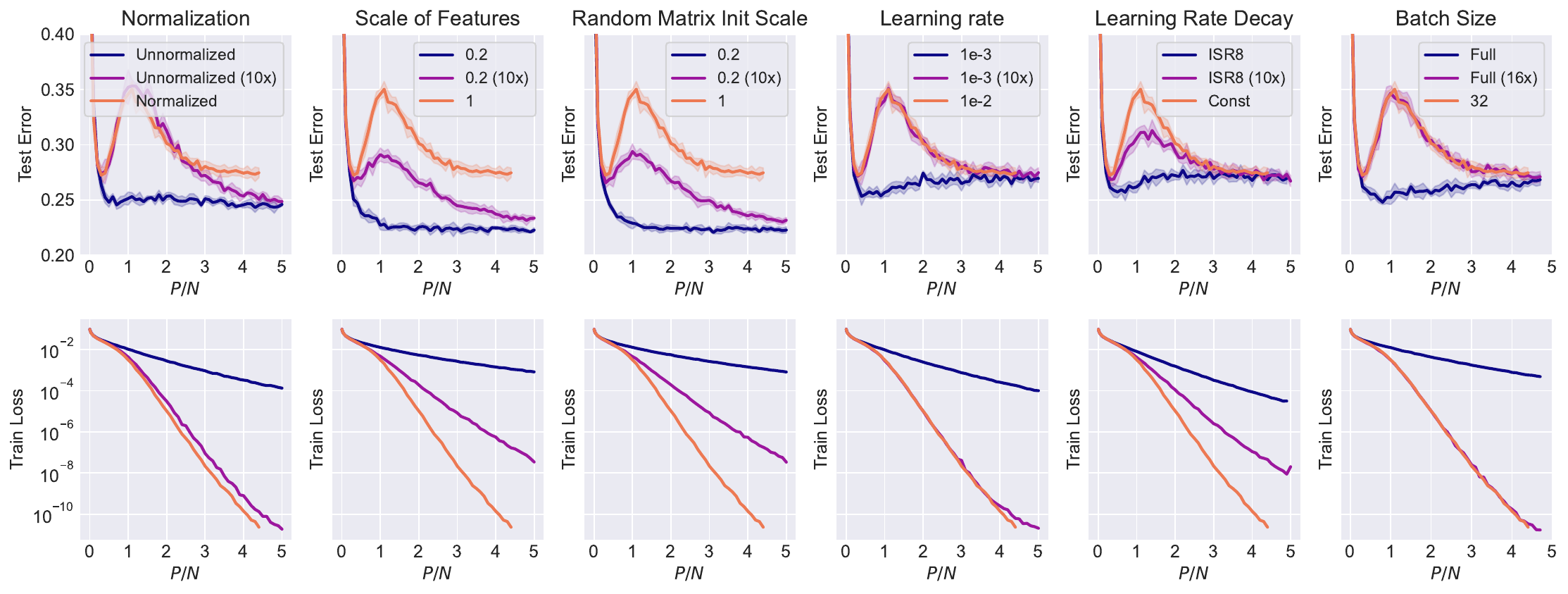}}
    \subfigure[RFM on CIFAR-10]{\includegraphics[width=\textwidth]{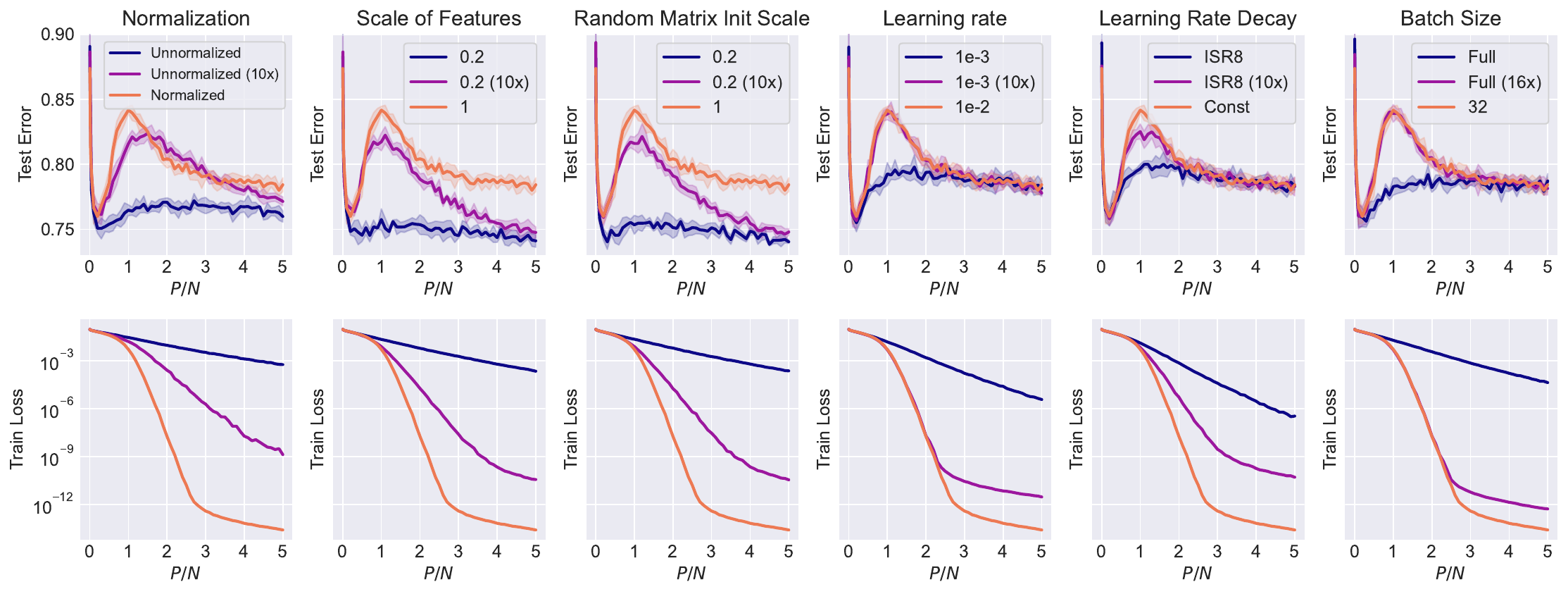}}
    \caption{Additional results to support \cref{fig:mnist_rfn_poor_optimizers_10x}. Test error and training loss curves of slow-convergence settings with 10x iterations of RFMs on Fashion-MNIST and CIFAR-10. We are able to recover the peaking phenomenon in all cases by scaling the number of iterations by a factor of 10.}
\end{figure}

\begin{figure}[H]
    \centering
    \subfigure[Two-layer neural network on MNIST]{\includegraphics[width=\textwidth]{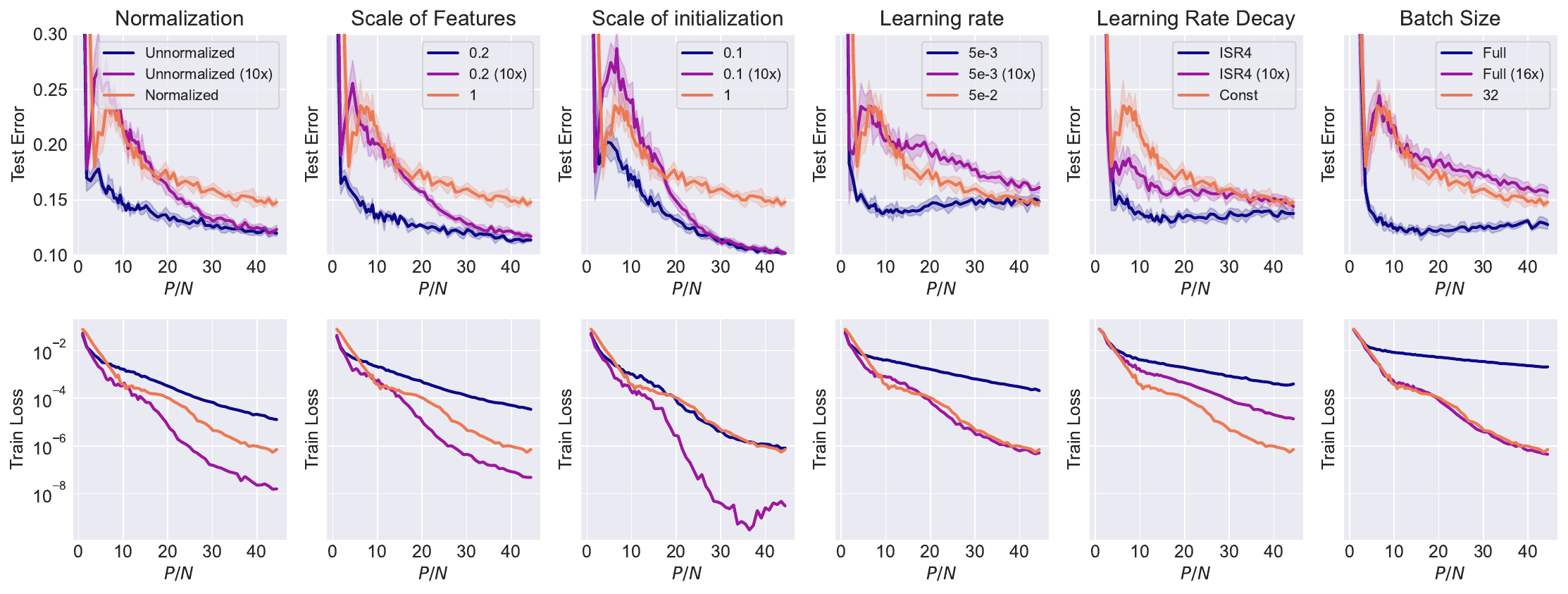}}
    \subfigure[Two-layer neural network on Fashion-MNIST]{\includegraphics[width=\textwidth]{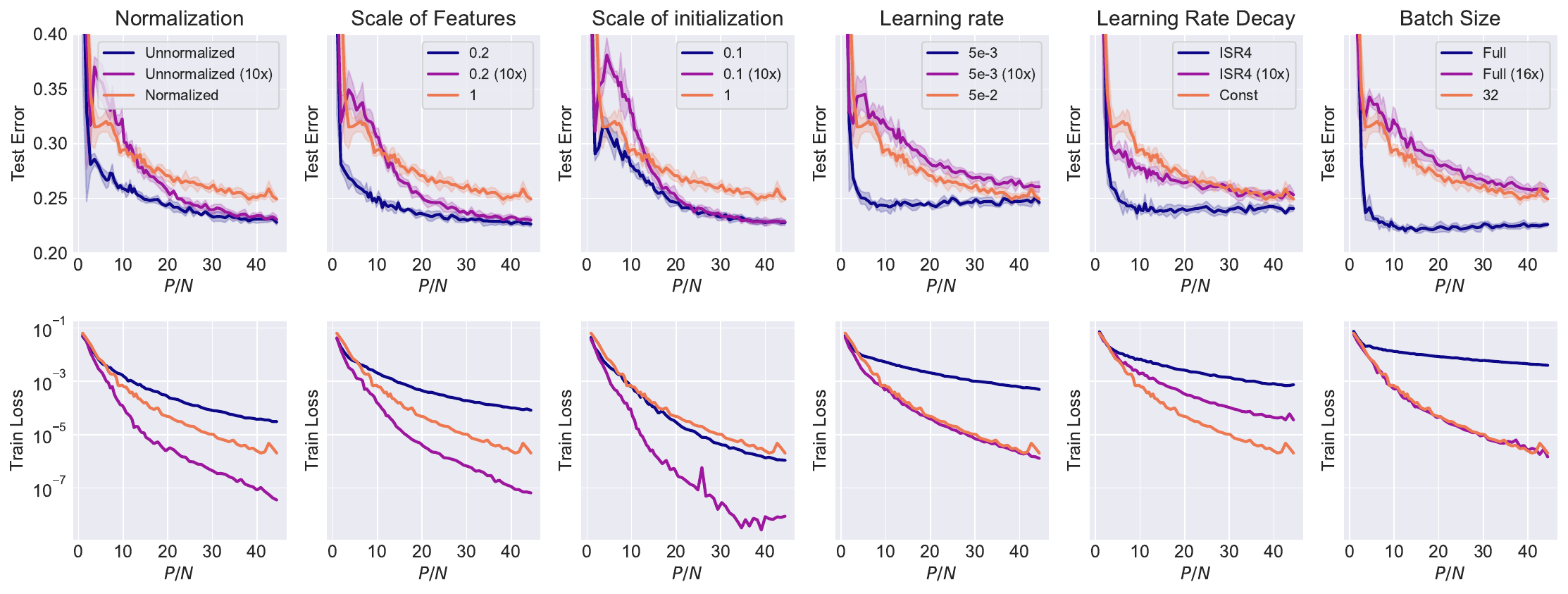}}
    \caption{Additional results to support \cref{fig:mnist_rfn_poor_optimizers_10x}. Test error and training loss curves of slow-convergence settings with 10x iterations of two-layer neural networks on MNIST and Fashion-MNIST. We are able to recover the peaking phenomenon in all cases by scaling the number of iterations by a factor of 10.}
\end{figure}

\end{document}